\documentclass[runningheads]{llncs}

\usepackage[framemethod=TikZ]{mdframed}
\mdfsetup{nobreak=false}

\usepackage{xspace}
\usepackage{colortbl}
\usepackage{amsmath,amsfonts,amssymb}
\usepackage{algorithm}
\usepackage{algpseudocode}
\usepackage{graphicx}
\usepackage{textcomp}
\usepackage{xcolor}
\usepackage{tikz}
\usepackage[most]{tcolorbox}
\usepackage{graphbox}
\usepackage{mathtools}
\usepackage{fancybox}
\usepackage{makecell}
\usepackage{multirow}
\usepackage{booktabs}
\usepackage{hyperref}
\hypersetup{hidelinks}
\usepackage{cleveref}
\usepackage[normalem]{ulem} \usepackage{listings}
\usepackage{caption}
\usepackage{subcaption}
\usepackage{longtable}
\usepackage{pifont}\usepackage{enumitem}
\usepackage{subfloat}
\usepackage[latin1]{inputenx}
\usepackage[T1]{fontenc}
\usepackage{soul}
\usepackage{courier}
\usepackage{svg}
\usepackage{lineno}

\newboolean{showcomments}
\setboolean{showcomments}{false}
\ifthenelse{\boolean{showcomments}}
{\newcommand{\nb}[2]{
  \fcolorbox{black}{yellow}{\bfseries\sffamily\scriptsize#1}
  {\sf\small$\blacktriangleright$\textit{#2}$\blacktriangleleft$}
 }
 
}
{\newcommand{\nb}[2]{}
 
}

\newcommand\federico[1]{\nb{federico}{#1}}

\newcommand{\Mnist}{MNIST\xspace}
\newcommand{\Taxinet}{TaxiNet\xspace}
\newcommand{\Yolo}{YOLOv4\xspace}
\newcommand{\fga}{FGA\xspace}
\newcommand{\Lymphoma}{LSC\xspace}

\lstdefinestyle{mystyle}{
    commentstyle=\color{gray},
    keywordstyle=\bfseries,
    numberstyle=\tiny\color{darkgray},
    basicstyle=\linespread{1.1}\ttfamily\footnotesize,
    breakatwhitespace=false,
    breaklines=true,
    captionpos=b,
    keepspaces=true,
    numbers=left,
    numbersep=10pt,
    showspaces=false,
    showstringspaces=false,
    showtabs=false,
    tabsize=4,
    frame=single,
}

\usepackage[framemethod=TikZ]{mdframed}
\newenvironment{Answer}[1][]{\ifstrempty{#1}{\mdfsetup{frametitle={\tikz[baseline=(current bounding box.east),outer sep=0pt]
      \node[line width=0pt,anchor=east,rectangle,draw=white,fill=white]
    ;}}
  }{\mdfsetup{frametitle={\tikz[baseline=(current bounding box.east),outer sep=0pt]
      \node[anchor=east,rectangle,draw=white,fill=white]
    {\strut #1};}}}\mdfsetup{innertopmargin=-5pt,linecolor=black,linewidth=0.5pt,topline=true,frametitleaboveskip=\dimexpr-\ht\strutbox\relax,}
  \begin{mdframed}[]\relax }{\end{mdframed}}
\newcommand\phase[1]{\tikz[baseline=(X.base)]\node [draw=black,fill=white,thick,rectangle,inner sep=2pt, rounded corners=2pt](X){\color{black}\textbf{#1}};}

\def\BibTeX{{\rm B\kern-.05em{\sc i\kern-.025em b}\kern-.08em
    T\kern-.1667em\lower.7ex\hbox{E}\kern-.125emX}}

\begin{document}
\title{Feature-Guided Analysis of Neural Networks:\\ A Replication Study}

\author{
    \href{https://orcid.org/0000-0002-3033-7371}{Federico Formica}\inst{1,*} \and
    \href{https://orcid.org/0009-0008-8494-0329}{Stefano Gregis}\inst{2,*} \and
    \href{https://orcid.org/0009-0008-0655-8335}{Aurora Francesca Zanenga}\inst{2}\and
    \href{https://orcid.org/0009-0008-1648-4130}{Andrea Rota}\inst{2}\and
    \href{https://orcid.org/0000-0003-3161-2176}{Mark Lawford}\inst{1} \and
    \href{https://orcid.org/0000-0001-5303-8481}{Claudio Menghi}\inst{2,1}
}

\institute{
    McMaster University, Hamilton, Canada 
    \email{\{formicaf, lawford\}@mcmaster.ca} \and
    University of Bergamo, Bergamo, Italy
    \email{\{s.gregis4, a.rota51\}@studenti.unibg.it, \{aurora.zanenga, claudio.menghi\}@unibg.it}\\
    * Both authors contributed equally to the paper
}

\authorrunning{Anonymous Author(s)}

\maketitle              \begin{abstract}
    
Understanding why neural networks make certain decisions is pivotal for their use in safety-critical applications.
Feature-Guided Analysis (\fga) extracts slices of neural networks relevant to their tasks. 
Existing feature-guided approaches typically monitor the activation of the neural network neurons to extract the relevant rules.
Preliminary results are encouraging and demonstrate the feasibility of this solution by assessing the precision and recall of Feature-Guided Analysis on two pilot case studies. 
However, the applicability in industrial contexts needs additional empirical evidence.

To mitigate this need, this paper assesses the applicability of \fga on a benchmark made by the \Mnist and \Lymphoma datasets.
We assessed the effectiveness of \fga in computing rules that explain the behavior of the neural network.
Our results show that \fga has a higher precision on our benchmark than the results from the literature.
We also evaluated how the selection of the neural network architecture, training, and feature selection affect the effectiveness of \fga.
Our results show that the selection significantly affects the recall of \fga, while it has a negligible impact on its precision.

\textbf{Submission type:} Empirical evaluation paper. %
\end{abstract}

\section{Introduction}
\label{sec:intro}
Deep Neural Networks (DNN) are widely used to support software engineering tasks and activities (e.g.,~\cite{boujida2024neural,khan2022software}). 
For example, neural networks have been extensively used to identify whether some input data belongs to a class or not.
Unlike traditional software, whose behavior is defined by engineers, the behavior of a neural network is learned from data~\cite{1634649}.
For example, given a set of images annotated with their corresponding classes, a neural network learns to classify the images based on the available classes. 
The behavior of a neural network is not explicitly defined by engineers, but learned from data, hampers the interpretation of the reasoning employed by neural networks~\cite{baier2019challenges}. 
For example, it can be difficult to determine how the neural network decides whether a certain image belongs to a class or not.
Therefore, the research community is investing significant effort to develop techniques that help engineers interpret the results and the actions selected by the neural network~\cite{molnar2022interpretable}.

Feature-Guided Analysis (\fga)~\cite{Gopinath_2023} extracts rules related to the most relevant neurons of a neural network, detecting the presence (or absence) of some features.
A rule has the form \textbf{pre}~$\rightarrow$~\textbf{post}, where the precondition (\textbf{pre})  is an assertion on the values assumed by (some of) the neurons of the neural network, 
and the postcondition (\textbf{post}) is an assertion on the presence (or absence) of a feature.
The rules extracted by FGA can support standard software engineering activities, such as testing, debugging, and requirements analysis.
For example, these rules can evaluate the quality of the datasets, retrieve and label new data, and understand scenarios where models make correct and incorrect predictions.

\fga was evaluated on the \Taxinet \cite{Beland_2020,Frew_2004} and the \Yolo-Tiny \cite{caesar2020nuscenes} benchmarks.
\Taxinet concerns center line tracking of airport runways,
\Yolo-Tiny (henceforth referred to as \Yolo) was used as an object detection model for autonomous driving.
For \Taxinet and \Yolo, the authors considered a dataset of 450 and 4000 images annotated with 12 features and 8 features.
The results show that \fga can extract rules involving a small number of neurons (compared to those of the neural network). 
Assessing the rules on a set of test images showed encouraging preliminary results~\cite{Gopinath_2023}: An acceptable precision and recall of the rules in the classification task.

Moving from research experiments to industrial systems requires significant efforts and activities~\cite{abadeer2021machine,8498185,breck2016s}. 
Repeating, reproducing, and replicating the experiments~\cite{cruz2019replication,juristo2012replication,shepperd2018role} are three of these activities highly relevant for ML techniques~\cite{10.1145/3477535,10.1145/3505243} and technology transfer~\cite{cockburn2020threats,daun2023industry,mendez2020open,shepperd2018replication,shepperd2018role}, and encouraged by the research community~\cite{EMSE}.

This paper replicates the experiments reported in the publication presenting Feature-Guided Analysis~\cite{Gopinath_2023}.
According to ACM~\cite{RRR} terminology, it is a replication study: It is conducted by a different team with a different experimental setup.
We considered a different experimental setup since the \Taxinet benchmark could not be made available by the authors of the original publication (it is a proprietary benchmark from Boeing). For the \Yolo benchmark, the neural network was retrained by the authors of the original publication; the retrained version was not shared, and the hyperparameters used for training were not provided.
These considerations hampered the reproduction of the original experiments.
Therefore, in this paper, we decided to consider the \Mnist (Modified National Institute of Standards and Technology database)~\cite{lecun1998} and \Lymphoma (Lymphoma Subtype Classification)~\cite{janowczyk2016} datasets for the following reasons.
They are commonly used for assessing ML solutions (e.g.,~\cite{9176802,xhaferra2022classification}), there are many publicly available neural networks trained for these datasets, retraining a neural network is possible with reasonable resources (and in practical time), and they are large and significant.
Therefore, they can provide significant and reliable results.

Our replication study assesses the generalizability of \fga to extract feature rules from DNN internals by extending its application to new DNN tasks and datasets. 
The goal is to analyze whether the precision and recall of the rules are comparable to those reported in the original publication~\cite{Gopinath_2023}.
The dataset and DNNs for \Mnist and \Lymphoma differ significantly from those considered in the original work.
We reimplemented the \fga algorithm since the code from~\cite{Gopinath_2023} is not publicly available. 

In our benchmark, \fga shows higher precision and a negligible reduction or better recall compared to the results from the research literature.
We also assessed how the selection of the neural network, training, and feature selection affect the effectiveness of the rules computed by \fga. 
Our results show that their selection does not significantly affect the test precision of the rules computed by FGA.
Conversely, it significantly influences their recall.

To summarize, our contributions are as follows:
\begin{itemize}
    \item A reimplementation of the \fga algorithm;
    \item The replication of the experiments from \fga on a new benchmark (consisting of two datasets \Mnist and \Lymphoma) and a systematic comparison with the results reported in the original publication;
    \item An empirical analysis on how the selection of the neural network, training, and feature selection affect the effectiveness of the rules computed by \fga;
    \item A systematic discussion on our results and their threats to validity;
    \item A replication package containing our dataset and implementation of \fga.
\end{itemize}

This paper is organized as follows.
\Cref{sec:background} summarizes FGA.
\Cref{sec:benchmark} and \Cref{sec:implementation}  describe our benchmark and implementation.
\Cref{sec:eval} presents our replication study.
\Cref{sec:discussion} discusses our results.
\Cref{sec:related} summarizes related work.
\Cref{sec:conclusion} concludes our work. 
 \section{Feature-Guided Analysis}
\label{sec:background}

A feedforward DNN is organized in multiple layers containing neurons.
Neurons are computational units that calculate their outputs based on the outputs of the neurons of the previous layer.
A trained DNN produces the likelihood that the input data belongs to a class depending on the calculations made by its neurons.

\begin{figure}[t]
    \centering
    \tikzstyle{output} = [coordinate]
\begin{tikzpicture}[auto,
 block/.style ={rectangle, draw=black, thick, fill=white!20, text width=8em,align=center, rounded corners},
 block1/.style ={rectangle, draw=blue, thick, fill=blue!20, text width=5em,align=center, rounded corners, minimum height=2em},
 line/.style ={draw, thick, -latex',shorten >=2pt},
 cloud/.style ={draw=red, thick, ellipse,fill=red!20,
 minimum height=1em}]

\node [output] (INITNode) {};
\draw (0,0) node[block,right of=INITNode, node distance=3cm] (B1) {\phase{1}  Neural Network Analysis};
\node [output, above of=B1, node distance=1cm] (NN) {};
\node [output, right of=B1, node distance=5cm] (ENDNode) {};
\node [block, right of=B1, node distance=6.2cm] (B2) {\phase{2} Decision Tree Computation};
\node [output, right of=B2, node distance=3cm] (ENDNodeTwo) {};

\draw[-stealth] (NN.south) -- (B1)
    node[midway,above]{\texttt{Model}, \texttt{Layers}};
\draw[-stealth] (B1.east) -- (B2.west)
    node[midway,align=left]{$\langle$\,\texttt{Neurons Activation},\\\quad \texttt{Feature Presence}\,$\rangle$};
\draw[-stealth] (INITNode.east) -- (B1.west)
    node[midway,above]{\parbox{1.5cm}{\texttt{Dataset},  \texttt{Features}}};
\draw[-stealth] (B2.east) -- (ENDNodeTwo.west)
    node[midway,above]{\parbox{1.5cm}{\texttt{Decision\\ Rules}}};

 \end{tikzpicture}     \caption{Feature-Guided Analysis: an Overview.}
    \label{fig:approach}
\end{figure}
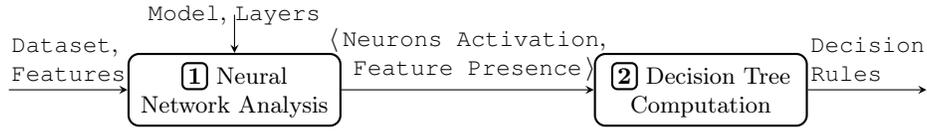

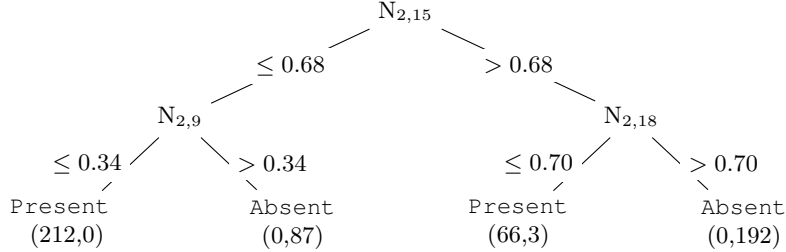
\begin{figure}[t]
    \centering
    \begin{tikzpicture}[level 1/.style={sibling distance=60mm},level 2/.style={sibling distance=30mm},level 3/.style={sibling distance=25mm}, level distance=40pt
]
\node[]{$\text{N}_{2,15}$}
child {
    node[] {$\text{N}_{2,9}$}
        child{
            node[align=center]{\texttt{Present } \\ (212,0)} edge from parent node[fill=white,left] {$\leq 0.34$}
        }
        child{
            node[align=center]{\texttt{Absent} \\ (0,87)} edge from parent node[fill=white,right] {$> 0.34$}
        }
        edge from parent node[fill=white] {$\leq 0.68$}
}
child { 
         node[align=center]{$\text{N}_{2,18}$}
         child{
                node[align = center]{\texttt{Present} \\ (66,3)} edge from parent node[fill=white,left] {$\leq 0.70$}
            }
          child{
            node[align=center]{\texttt{Absent} \\ (0,192)} edge from parent node[fill=white,right] {$> 0.70$}
          }
            edge from parent node[fill=white] {$> 0.68$}
    };
\end{tikzpicture}
      \caption{Example of decision tree structures like the ones from \fga.}
    \label{fig:decisionTree}
\end{figure}

\Cref{fig:approach} provides an overview of \fga.
\fga takes as input a \emph{trained} feedforward neural network (\texttt{Model}) and a set of layers (\texttt{Layers}) to be considered by FGA for the computation of the rules.
The algorithm considers a dataset (\texttt{Dataset}) and a set of features (\texttt{Features}) and analyzes which neurons are activated (\texttt{Neurons Activation}) when the feature is present/absent (\texttt{Feature Presence}) while the neural network processes the images from the dataset searching for these features (\phase{1}).
The activation values of the different neurons and the labels indicating the presence or absence of a feature are used to compute a decision tree based on the neuron activation values~(\phase{2}). 
Decision rules (\texttt{Decision Rules}) are extracted from the tree considering a complete path from the root node to a leaf.
\Cref{fig:decisionTree} shows an example of a decision tree computed by  \fga. 
The decision tree has three layers, one root node, and four leaf nodes. 
Each node is a neuron $\text{N}_{a,b}$ where $a$ is the layer number and $b$ is the neuron number. 
For example, the node $\text{N}_{2,15}$ is associated with the neuron $15$ from the second layer.
Each branch is a condition on the activation value of that node.
For example, the branch from $\text{N}_{2,15}$ to $\text{N}_{2,9}$ considers activation values for the node $\text{N}_{2,15}$ lower than $\leq 0.68$.
Each leaf node is associated with a label that indicates the presence or absence of a feature. 
Intuitively, considering the neuron activation values and the corresponding predictions for the feature presence or absence made by the neural network, the decision tree specifies whether it is more likely for the feature to be present or absent.
For example, according to the decision tree in \Cref{fig:decisionTree}, when the activation values for the neurons $\text{N}_{2,15}$ and $\text{N}_{2,9}$ are lower than or equal to $0.68$ and $0.34$, the feature is likely to be present.
The node is also associated with a tuple of values (e.g., ``$(212,0)$''): The first value defines the number of input images from the dataset with the feature specified by the node, and the second value is the number of inputs without the feature. 
\fga considers only pure leaves (i.e., leaf nodes that contain only inputs of the correct class), which means that \texttt{Present} (66,3) would not be considered for the creation of a rule. For example, the rule extracted from the leftmost leaf node would be $(\text{N}_{2,15} \leq 0.68 \land \text{N}_{2,9} \leq 0.34) \rightarrow \texttt{Present}$.

 \section{Benchmark}
\label{sec:benchmark}
Our benchmark consists of the MNIST (Modified National Institute of Standards and Technology database)~\cite{lecun1998} and \Lymphoma (Lymphoma Subtype Classification)~\cite{janowczyk2016} datasets.

\Mnist is a dataset of handwritten gray-scale images representing digits. 
Each image is associated with a label indicating the corresponding digit.
The dataset contains 70'000 images: 60'000 images representing the training dataset, and 10'000 images representing the test dataset.

\Lymphoma is a dataset from the National Institute on Aging. 
It contains a collection of histopathological images for the classification of three lymphoma types: Chronic Lymphocytic Leukemia (CLL), Follicular Lymphoma (FL), and Mantle Cell Lymphoma (MCL).
The dataset contains 374 images (113 CLL, 139 FL, and 122 MCL).
Following the approach of the original paper, each image ($1388\times1040$ px) was cropped into 1376 overlapping patches of $36\times36$ px with a stride of 32, yielding a total of 514'624 patches.
The original work proposed a winner-take-all decision logic where the DNN returned the classification for each patch, and the most frequent class became the classification of the entire image.
The original dataset provides the ground-truth classification for the original 374 images, and not for the individual patches, since some patches may not contain evidence of any lymphoma type.
To solve this problem, we decided to filter the dataset of 514'624 patches and considered only the patches for which the network proposed in~\cite{janowczyk2016} returned a classification score above $95\%$.
Since this network achieves a high classification accuracy (96.58\% $\pm$ 0.01\%~\cite{janowczyk2016}), this ensures that only the patches that can be confidently classified in one of the three classes are considered.
This filtering process reduced the dataset to 442'398 patches, out of which 135'574 were classified as CLL, 169'367 as FL, and 137'457 as MCL.

 \section{Implementation}
\label{sec:implementation}
We could not directly reuse the code from~\cite{Gopinath_2023} since it is not publicly available. 
Therefore, we decided to reimplement \fga.
Our source code is designed to work with arbitrary network architectures trained on the \Mnist and \Lymphoma benchmarks and can support feature sets where a single image is tagged with multiple features.
This functionality enables us to consider rules that detect digits consisting of circles, i.e., data labeled with a ``0'', ``6'', ``8'', or ``9'', while at the same time detecting the presence of individual digits, i.e., data labeled only with a ``0''.
Our code computes a separate decision tree for every feature from a dataset containing labels indicating whether a feature is present, absent, or the input is misclassified.
We then remove the inputs misclassified by the neural network from the training dataset before building the decision tree.\footnote{For MNIST networks, have a high accuracy, so the misclassified inputs are a small minority (at most 4\% for M-DNN1) and they do not affect the results significantly.}

We implemented \fga by adapting Prophecy~\cite{Gopinath_2019,gopinath2025prophecy}, a tool that (unlike \fga) (a)~computes rules based on the activation status ``on''/``off'' of the neurons of the neural network and (b)~extracts rules for correct vs missclassified inputs.
The code was written to work with DNN implemented using the TensorFlow Python module. 
We considered TensorFlow 2.13.
Our implementation is available as part of the Replication package~\cite{replication}.

 \section{Evaluation}\label{sec:eval}
Our replication study considers the following research question:
\begin{enumerate}[start=1,leftmargin=1cm,label={\bfseries RQ\arabic*:}]
    \item How effective is \fga on our benchmark compared with the results reported in the original work proposing \fga? (\Cref{sec:replicability})
\end{enumerate}
This research question assesses how the effectiveness of \fga on our benchmark compares to the results originally reported by the authors~\cite{Gopinath_2023}. 
For this reason, this experiment replicates as closely as possible the configuration used for the experiments of their original work.

Additionally, our replication study evaluated how the training, type, and feature selection of the neural networks affect the effectiveness of \fga. Specifically, we consider the following additional research questions.

\begin{enumerate}[start=2,leftmargin=1cm,label={\bfseries RQ\arabic*:}]
    \item How does the selection of the neural network affect the effectiveness of \fga? (\Cref{sec:nncomparison})
    \item How does the composition of the training dataset affect the effectiveness of \fga? (\Cref{sec:training}) 
\item How does the feature selection affect the effectiveness of \fga? (\Cref{sec:featureSelection})
\end{enumerate}

\subsection{Replicability (RQ1)} 
\label{sec:replicability}
To assess how replicable the experiments from the original work proposing \fga are, we apply \fga to a neural network trained on the \Mnist dataset and one trained on the \Lymphoma dataset, and compare the rules we obtain with those from the original paper. 

\emph{Study Subject.} 
For the \Mnist dataset, we consider the neural network architecture (M-DNN1) proposed by the authors of Prophecy \cite{Gopinath_2019}, but not included in the original publication. The network has $10$ layers, with $2$ convolutional and $2$ dense layers.
The activations of the convolutional and dense layers are considered as separate layers.
We did not train the neural network for this experiment, but we downloaded a pretrained version.
For the \Lymphoma dataset, we utilized the architecture L-DNN1~\cite{janowczyk2016}.
The 12-layer CNN begins with a feature extractor composed of three sequential Convolution-ReLU-Pooling blocks, then the classification section consists of two fully-connected layers, and it terminates with a \texttt{SoftmaxWithLoss} layer to compute the class probabilities.
We downloaded the pretrained Caffe model~\cite{lymphomacaffe2016} for the first 5-fold cross-validation split and converted it to the ONNX for use in our framework.

\begin{table*}[t]
    \centering
    \caption{Feature, set of digits, and explanation for the \Mnist dataset.}
    \begin{tabular}{p{1.8cm} p{2.5cm} l }
        \toprule
        \textbf{Feature}    &\textbf{Set of digits} & \textbf{Explanation} \\
        \midrule
         Digit 0 & 0 & Represents the digit ``0''.\\
         \quad$\vdots$ & $\vdots\,$\\
         Digit 9 & 9 & Represents the digit ``9''.\\
         2 and 7 & 2,7 & Represents two digit graphically similar.\\
         9 and 6 & 9,6 & Represents two digit graphically similar.\\
         Line & 1,4,7 & Represents three digit sharing a commonality (line).\\
         Circle & 0,6,8,9 & Represents four digit sharing a commonality (circle).\\
        \bottomrule
    \end{tabular}
    \label{tab:features}
\end{table*}

\emph{Methodology.} We replicated the methodology reported in~\cite{Gopinath_2023}. 
For \Mnist, we select the only hidden dense layers of the network (i.e., the neurons from the first dense layer, after the activation of the first dense layer, and before the activation of the last dense layer) to be considered for extracting the \fga rules.
We considered the features from \Cref{tab:features}.
These features include the presence of single digits ``0'', ``1'', $\ldots$, ``9'', couples of digits graphically similar ``2'' and ``7'', and ``9'' and ``6'', sets of digits sharing a similar characteristics such as the presence of a straight line (``1'', ``4'', and ``7'') or a circle (``0'', ``6'', ``8'', and ``9'').

For \Lymphoma, we select the neurons from two dense layers to be considered for extracting the \fga rules, as these are the only dense layers in the network.
For the features, we considered the three individual classes ``CLL'', ``FL'', and ``MCL'', and their pairwise combinations ``CLL \& FL'', ``CLL \& MCL'', and ``FL \& MCL''.

We run \fga. 
We extracted all the rules from the decision trees related to pure nodes (i.e., rules with  100\% train precision). 
We selected the rule associated with the leaf node with the highest number of input samples for each feature. 
Then, we computed the train and test metrics related to this rule.

We computed two metrics: precision ($\frac{TP}{TP + FP}$) and recall ($\frac{TP}{TP + FN}$).
A True Positive (TP) is an input that displays the feature and satisfies the precondition of the rule, a False Positive (FP) is an input that satisfies the precondition of the rule but does not show the feature, and a False Negative (FN) is an input that displays the feature but does not satisfy the precondition of the rule.

The best rule for each feature was computed and evaluated at all the selected layers.
For every feature, we selected the best-performing (in terms of recall) rule among the ones extracted at the specified layers.

\emph{Results.} 
\Cref{tab:rq1comparison} reports the recall ($R_{tr}$, $R_{te}$) of the rules on the training ($tr$) and the test ($te$) dataset, their precision on the test dataset ($P_{te}$), and the length ($Len$) of the extracted rules for the \Mnist (\Cref{tab:rq1-mnist}), \Lymphoma (\Cref{tab:rq1-lymphoma}), \Taxinet (\Cref{tab:rq1-taxinet}), and \Yolo (\Cref{tab:rq1-yolo})  datasets. 
For \Mnist and \Lymphoma datasets, we also reported the best rule for each feature. 
Considering the length of the rules, only the first and last clauses are reported, but the complete rules can be found in the replication package.
The results associated with \Taxinet and \Yolo are from their original publication \cite{Gopinath_2023}.
The precision of the training dataset is omitted since the rules are associated with pure nodes (and therefore have 100\% train precision).
The last row of the tables from \Cref{tab:rq1-mnist,tab:rq1-lymphoma,tab:rq1-taxinet,tab:rq1-yolo} report the average recall ($R_{tr}$, $R_{te}$) on the training ($tr$) and the test ($te$) dataset, precision on the test dataset ($P_{te}$), and the length ($Len$) of the extracted rules. 
The maximum and minimum absolute values of recall, precision, and length ($Len$) for the features are reported with a blue and orange background.

\begin{table*}[!t]
\centering
\caption{
Train ($R_{tr}$) and Test ($R_{te}$) Recall, Test Precision ($P_{te}$), length (Len), and portion of the rules of M-DNN1 (MNIST) and L-DNN1 (LSC). 
}
\label{tab:rq1comparison}
\subfloat[\Mnist.]{
\scriptsize
\begin{tabular}{l | r r r r l}
        \toprule
         \textbf{Feature}    & $\mathbf{R_{tr}}$    & $\mathbf{P_{te}}$  & $\mathbf{R_{te}}$ & \textbf{Len} & \textbf{Rule}\\
        \midrule
        Digit 0     & 82.69    & \cellcolor{blue!25}100.00    & 85.80 & 22 & $(N_{1,116} > 16.14 \land \dots \land N_{1,42} > -6.06)\Rightarrow \{0\}$ \\
        Digit 1     & \cellcolor{blue!25} 89.47    & 99.90    & \cellcolor{blue!25} 90.17 & 15 & $(N_{3,1} > 4.34 \land \dots \land N_{3,1} > 5.34)\Rightarrow \{1\}$ \\
        Digit 2     & 66.46    & 99.57    & 69.62 & 12 & $(N_{3,2} > 6.21 \land \dots \land N_{3,8} \leq 5.78)\Rightarrow \{2\}$ \\
        Digit 3     & 65.99    & 99.69    & 67.36 & 16 & $(N_{3,3} > 7.00 \land \dots \land N_{3,2} > -1.02)\Rightarrow \{3\}$ \\
        Digit 4     & 76.15    & 99.60    & 77.74 & 22 & $(N_{3,4} > 6.38 \land \dots \land N_{3,5} > -6.49)\Rightarrow \{4\}$ \\
        Digit 5     & 71.84    & \cellcolor{blue!25}100.00    & 73.50& 12 & $(N_{3,5} > 6.68 \land \dots \land N_{3,9} > -5.07)\Rightarrow \{5\}$ \\
        Digit 6     & 76.99    & 99.58    & 76.66 & 33 & $(N_{1,19} > 5.66 \land \dots \land N_{1,4} \leq 2.54)\Rightarrow \{6\}$ \\
        Digit 7     & 49.93    & 99.81    & 52.87 & 27 & $(N_{1,86} > 14.47 \land \dots \land N_{1,26} \leq 1.89)\Rightarrow \{7\}$ \\
        Digit 8     & 59.08    & 99.81    & 61.63 & 18 & $(N_{3,8} > 4.54 \land \dots \land N_{3,9} \leq 3.61)\Rightarrow \{8\}$ \\
        Digit 9     & 57.59    & 99.31    & 60.23 & 33 & $(N_{1,65} > 3.02 \land \dots \land N_{1,25} \leq 13.54)\Rightarrow \{9\}$ \\
        2 and 7     & 32.53   & \cellcolor{orange!25} 98.66    & 33.42 & 42 & $(N_{1,91} > 3.06 \land \dots \land N_{1,118} > -9.96)\Rightarrow \{2,7\}$ \\
        9 and 6     & 36.38    & 98.85    & 36.27 & 14 & $(N_{3,6} > 5.59 \land \dots \land N_{3,5} \leq 4.06)\Rightarrow \{6,9\}$ \\
        Line     & \cellcolor{orange!25} 28.53    & \cellcolor{blue!25} 100.00    & \cellcolor{orange!25} 28.73 & \cellcolor{orange!25} 10 & $(N_{3,7} > -3.81 \land \dots \land N_{3,4} > -4.24)\Rightarrow \{1,4,7\}$ \\
        Circle    & 28.60    & 99.03    & 30.16 & \cellcolor{blue!25} 52 & $(N_{1,8} > 3.11 \land \dots \land N_{1,59} \leq -8.20)\Rightarrow \{0,6,8,9\}$ \\
        \midrule
        Average    & 58.73    & 99.63    & 60.30 & 23.4 & \\
        \bottomrule
\end{tabular}
\label{tab:rq1-mnist}
}\\
\subfloat[\Lymphoma.]{
\scriptsize
\begin{tabular}{l | r r r r l}
    \toprule
    \textbf{Feature} & $\mathbf{R_{tr}}$ & $\mathbf{P_{te}}$ & $\mathbf{R_{te}}$ & \textbf{Len} & \textbf{Rule}\\
    \midrule
    CLL & \cellcolor{blue!25} 66.77 & 99.12 & \cellcolor{blue!25} 63.75 & \cellcolor{blue!25} 22 & $(N_{1,6} > 0.26 \land \dots \land N_{1,1} \leq 0.06)\Rightarrow \{CLL\}$ \\
    FL & 62.98 & 99.07 & 58.06 & 13 & $(N_{1,41} \leq -0.24 \land \dots \land N_{1,12} \leq 0.07)\Rightarrow \{FL\}$ \\
    MCL & 60.41 & 99.94 & 60.46 & 16 & $(N_{1,52} > 0.44 \land \dots \land N_{1,26} \leq -0.17)\Rightarrow \{MCL\}$ \\
    CLL \& FL & 35.49 & \cellcolor{orange!25} 99.06 & 32.78 & \cellcolor{orange!25} 7 & $(N_{1,52} \leq 0.44 \land \dots \land N_{1,12} \leq 0.07)\Rightarrow \{CLL,FL\}$ \\
    MCL \& FL & 34.87 & 99.10 & 32.93 & 15 & $(N_{1,6} \leq 0.26 \land \dots \land N_{1,29} \leq -0.53)\Rightarrow \{MCL,FL\}$ \\
    CLL \& MCL & \cellcolor{orange!25} 28.93 & \cellcolor{blue!25} 99.95 & \cellcolor{orange!25} 28.50 & 11 & $(N_{1,41} > -0.24 \land \dots \land N_{1,26} \leq -0.17)\Rightarrow \{CLL,MCL\}$ \\
    \midrule
    Average & 48.24 & 99.37 & 46.08 & 14.0 & \\
    \bottomrule
\end{tabular}
\label{tab:rq1-lymphoma}
}\\
\subfloat[\Taxinet.]{
\scriptsize
\begin{tabular}{l | r r r r }
        \toprule
        \textbf{Feature}   & $\mathbf{R_{tr}}$    & $\mathbf{P_{te}}$  & $\mathbf{R_{te}}$ & \textbf{Len} \\
        \midrule
        Center-line: present    & 92.0    & 93.0    & \cellcolor{blue!25} 100.0 & 4 \\
        Center-line: absent  & \cellcolor{orange!25} 40.0    & \cellcolor{blue!25} 100.0    &\cellcolor{orange!25} 12.0 & \cellcolor{orange!25} 2  \\   
        Shadow: present & 86.0    & \cellcolor{blue!25} 100.0    & 69.2 & 3  \\
        Shadow: absent    & 94.5    & 97.0   & \cellcolor{blue!25} 100.0 & 3  \\
        Skid: dark    & 52.5    & 94.4    & 43.5 & \cellcolor{orange!25} 2  \\
        Skid: no  & 60.0    &  0.0    &  0.0  & \cellcolor{orange!25} 2 \\
        Skid: light    & \cellcolor{blue!25} 97.8    & 93.4    & 95.0  & 4  \\
        Position: right & 90.0    & 92.3    & 95.1 & 3  \\
        Position: left & 91.0    & \cellcolor{blue!25} 100.0    & 75.2  & 3 \\
        Position: on & 45.0    &\cellcolor{orange!25} 13.5    & 45.5 & 6 \\
        Heading: away & 65.0  & 62.2    & 90.6  & 3 \\
        Heading: towards & 83.0    & 73.9    & 16.5 & \cellcolor{blue!25} 7  \\
        \midrule
        Average    & 74.73    & 76.65    & 61.88  & 3.5   \\
        \bottomrule
\end{tabular}
\label{tab:rq1-taxinet}
}
\subfloat[\Yolo.]{
\scriptsize
    \begin{tabular}{l | r r r r}
       \toprule
        \textbf{Feature}   & $\mathbf{R_{tr}}$    & $\mathbf{P_{te}}$  & $\mathbf{R_{te}}$ & \textbf{Len} \\
        \midrule
        Pedestrian moving: present    & 48.0    & 72.0    & 29.0 & 21 \\
        Pedestrian moving: absent  & 40.0    & 74.0    & 29.0 & 15  \\   
        Vehicle parked: present & \cellcolor{orange!25} 25.0    & 71.0    &\cellcolor{orange!25} 20.0 & \cellcolor{orange!25} 10  \\
        Vehicle parked: absent    & 43.0  & 70.0  & 32.0 & \cellcolor{blue!25} 29 \\
        Pedestrian: present    & 57.0 & 70.0 & 35.0 & 25  \\
        Pedestrian: absent  & 41.0 & 77.0 & 22.0 & 14 \\
        Vehicle: present  & \cellcolor{blue!25} 75.0 & \cellcolor{blue!25} 91.0 & \cellcolor{blue!25} 59.0 & 20  \\
        Vehicle: absent & 50.0   & \cellcolor{orange!25} 69.0 & 31.0 & 11  \\
        \midrule
        Average    & 47.38 & 74.25 & 32.13  & 18.1 \\
        \bottomrule
    \end{tabular}
    \label{tab:rq1-yolo}
}
\end{table*}

\emph{Train Recall}. 
For \Mnist, the rules for features ``Digit 1'' and  ``presence of a straight line'' have the highest (89.47\%) and lowest (28.53\%) train recall.
For \Lymphoma, the rules for the ``CLL'' and ``CLL \& MCL'' features have the highest (66.77\%) and lowest train recall (28.93\%).
For the \Taxinet dataset, the rules for features ``Skid: light'' and ``Center-line: absent'' have the highest (97.8\%) and lowest (40.0\%) train recall. 
For the \Yolo network, the rules ``Vehicle: present'' and ``Vehicle parked: present'' have the highest (75.0\%) and lowest (25.0\%) train recall.
The average \fga recall on the training dataset for the \Mnist, \Lymphoma, \Taxinet, and \Yolo datasets is 58.73\%, 48.24\%, 74.73\%, and 47.38\%. 

Our results show that \fga is consistent with what is reported in the literature, as our highest train recall (89.47\%) is lower than the highest train recall (97.8\%) from the literature, and our lowest train recall (28.53\%) is higher than their lowest train recall (25.0\%) from the literature.
Our average train recall for \Mnist (58.73\%) is 16.00\% lower than \Taxinet (74.73\%), and 11.36\% higher than  \Yolo (47.38\%).
Our average train recall for \Lymphoma (48.24\%) is 26.49\% lower than \Taxinet (74.73\%), and 0.86\% higher than \Yolo  (47.38\%).

A comparison of the train recall for the different features of \Mnist shows that the train recall of features associated with multiple digits, i.e., two (``2'' and ``7'' and ``9'' and ``6''), three (Line) and four (Circle) digits, is lower than features representing a single digit.
Intuitively, the rules computed by \fga are more likely to miss their presence for features shared by multiple digits compared with features referring to single digits (i.e., the recall of features involving multiple digits is lower than features referring to single digits). Like \Mnist, for \Lymphoma, the rules for features representing a single class have a significantly higher recall than those for features combining multiple classes.

\emph{Test Precision.}
For \Mnist, the rules for the features ``Digit 0'', ``Digit 5'', and ``Line'' have the highest (100.0\%) test precision. 
The rule for the feature ``2 and 7'' has the lowest test precision (98.66\%). 
For \Lymphoma, the rule for the feature ``CLL \& MCL'' has the highest test precision (99.95\%), while ``CLL \& FL'' has the lowest (99.06\%).
For the \Taxinet dataset, the rules for the features ``Center-line: absent'', ``Shadow: present'', and ``Position: left'' have the highest (100.0\%) test precision.
The rule for the feature ``Skid: no'' has the lowest  (0.0\%) test precision. 
This result is justified since, according to the results reported by the authors in their paper \cite{Gopinath_2023}, for \Taxinet, only 5 images of the training dataset satisfied the rule, while for \Mnist and \Lymphoma, each feature is associated with approximately 6000 and 147466 instances from the training dataset, as it is balanced.
Therefore, for \Taxinet, it is likely that the test dataset contained only very few images showing this feature.
We will therefore exclude this feature from the comparison and consider ``Position: on'' as the feature with the lowest test precision (13.5\%).
For \Yolo, the rules for the features ``Vehicle: present'' and ``Vehicle: absent'' have the highest (91.0\%) and lowest (69.0\%) test precision.
This result shows that \fga has higher test precision for \Mnist and \Lymphoma than the one reported in the literature. 
The lowest test precision for \Mnist is 98.66\% and for \Lymphoma is 99.06\%, while for \Taxinet is 13.5\%, and for \Yolo is 69\%. 
The average test precision for \Mnist (99.63\%) is higher than the \Taxinet (+22.98\%) and \Yolo (+25.38\%) datasets. 
The average test precision for \Lymphoma (99.37\%) is higher than \Taxinet (+22.72\%) and \Yolo (+25.12\%) datasets.

\emph{Test Recall.}
For \Mnist, the rules for the features ``Digit 1'' and ``Line'' have the highest (90.17\%) and lowest (28.73\%) test recall.  
For \Lymphoma, the rules for the ``CLL'' and ``CLL \& MCL''  features achieve the highest (63.75\%) and lowest (28.50\%) test recall.
For \Taxinet, the rules for the features ``Center-line: present'' and ``Shadow: absent'' have the highest (100.0\%) test recall.
The rule for the feature ``Center-line: absent'' has the lowest test recall (12.0\%). 
For \Yolo, the rules ``Vehicle: present'' and ``Vehicle parked: present'' have the highest (59.0\%) and lowest (20.0\%) test recall.
The average test recall is 60.30\% for \Mnist, 46.08\% for \Lymphoma, 61.88\% for \Taxinet, and 32.13\% for \Yolo.
Our results confirm the findings from the literature: Our highest test recall (90.17\%) is lower than the maximum value from the literature (100.0\%), and our minimum test recall (28.50\%) is higher than the minimum value from the literature (12.0\%). 
The test recall for \Mnist (60.30\%) is lower than for \Taxinet (-1.58\%) and higher than for \Yolo (+28.18\%).
The test recall for \Lymphoma (46.08\%) is lower than for \Taxinet (-15.80\%) and higher than for \Yolo (+13.95\%).
Like the train recall, the rules associated with features aggregating multiple features perform worse than those for a single feature.

\emph{Number of pre-conditions}.
For \Mnist, the rules ``Line'' and ``Circle'' have the minimum (10) and the maximum (52) number of preconditions.
For \Lymphoma, the rules ``CLL \& FL'' and ``CLL'' have the minimum (7) and maximum (22) number of preconditions.
On average, rules from \Mnist, \Lymphoma, \Taxinet, and \Yolo have 23.4, 14.0, 3.5, and 18.1 conjunctive clauses in their precondition. 
Therefore, the rules obtained for the \Mnist dataset are longer than those of \Taxinet and \Yolo, while those obtained for the \Lymphoma dataset are longer than those of \Taxinet, but shorter than \Yolo. 
Despite being longer than those from the literature, the rules are still sufficiently compact (and therefore likely to be interpretable by engineers).

\begin{Answer}[RQ1 --- Replicability]
For \Mnist and \Lymphoma, our results show that \fga has a higher test precision than \Taxinet (+22.98\% and +22.72\%) and \Yolo (+25.38\% and +25.38\%), with a test recall that is lower than \Taxinet ($-1.60\%$ and $-15.80\%$), but higher than \Yolo  (+28.16\% and +13.95\%). 
\end{Answer} 

\subsection{Influence of Neural-Network Selection (RQ2)}

\label{sec:nncomparison}
To assess how the selection of the neural network influences the effectiveness of \fga, we compared the rules extracted by the neural network architectures from \Cref{sec:replicability} (M-DNN1 and L-DNN1) with three other architectures presented in the following.

\emph{Study Subjects}. 
For \Mnist, our study subjects are M-DNN1 and two other neural networks from the literature (M-DNN2~\cite{crabbe2022} and M-DNN3~\cite{lecun1998}). 
M-DNN2 has 12 layers organized in two convolutional layers, each followed by a dropout layer and a pooling layer, one flatten, two linear layers, each followed by a dropout, and one last linear layer on the output. The linear layers respectively have 10, 5, and 10 neurons. 
M-DNN3 (LeNet5) has a different size, number of kernels in the convolutional layers, and a bigger size of the dense layers than M-DNN1 and M-DNN2.
Specifically, M-DNN3 has eight layers: two convolutional layers, each followed by a pooling layer, one flatten layer followed by three dense layers, with 120, 84, and 10 neurons each.

For \Lymphoma, we use L-DNN1 and L-DNN2, an upgraded version of L-DNN1 with two ReLU-Dropout blocks after each of the two fully-connected layers.

\emph{Methodology}. We trained M-DNN2 and M-DNN3 on the \Mnist dataset using standard parameters compatible with both architectures~\cite{crabbe2022}.
For L-DNN2, we downloaded the pretrained Caffe model~\cite{lymphomacaffe2016} for the first 5-fold cross-validation split. We converted it to the ONNX for use in our framework.

We run \fga (as detailed in \Cref{sec:replicability}) using the features from \Cref{tab:features}.
For M-DNN2, we extract activations after four layers: The first linear layer and the subsequent dropout, the second linear layer, and the subsequent dropout.
For M-DNN3, we extract the rules after the first and second dense layers, as they are the two hidden dense layers of the network.
For L-DNN2, we extract the rules from the two dense layers, as these are the only dense layers in the network.

\begin{table*}[t!]
    \centering
    \caption{Train ($R_{tr}$) and Test ($R_{te}$) Recall and the Test Precision ($P_{te}$) of different neural networks.}
    \label{sec:TableComparison}
    \subfloat[M-DNN1, M-DNN2, and M-DNN3.]{
     \scriptsize
    \begin{tabular}{l | r r r | r r r | r r r}
        \toprule
        \multirow{2}{1.3cm}{\textbf{Feature}}   &\multicolumn{3}{c|}{\textbf{M-DNN1}}  &\multicolumn{3}{c|}{\textbf{M-DNN2}}  &\multicolumn{3}{c}{\textbf{M-DNN3}} \\
          \cmidrule{2-10}
            & $\mathbf{R_{tr}}$    & $\mathbf{P_{te}}$  & $\mathbf{R_{te}}$ & $\mathbf{R_{tr}}$    & $\mathbf{P_{te}}$    & $\mathbf{R_{te}}$ & $\mathbf{R_{tr}}$    & $\mathbf{P_{te}}$    & $\mathbf{R_{te}}$ \\
        \midrule
        Digit 0     & 82.69    & \cellcolor{blue!25} 100.00    & 85.80     & 96.04 & \cellcolor{orange!25} 99.46 & 94.56 & 85.44 & 99.42 & 88.10 \\
        Digit 1     & \cellcolor{blue!25} 89.47    & 99.90    & \cellcolor{blue!25} 90.17     & 88.81 & \cellcolor{blue!25} 100.00 & 90.37 & 91.07 & 99.53 & 92.74 \\
        Digit 2     & 66.46    & 99.57    & 69.62     & 92.16 & \cellcolor{blue!25} 100.00 & 91.82 & 89.95 & 99.79 & 90.46 \\
        Digit 3     & 65.99    & 99.69    & 67.36     & 79.18 & \cellcolor{blue!25} 100.00 & 81.12 & 70.01 & 99.14 & 70.74 \\
        Digit 4     & 76.15    & 99.60    & 77.74     & 68.51 & \cellcolor{blue!25} 100.00 & 71.68 & 90.94 & \cellcolor{blue!25} 100.00 & 92.88 \\
        Digit 5     & 71.84    & \cellcolor{blue!25} 100.00    & 73.50     & 83.80 & 99.73 & 84.85 & 83.88 & \cellcolor{orange!25} 99.08 & 84.94 \\
        Digit 6     & 76.99    & 99.58    & 76.66     & 84.82 & \cellcolor{blue!25} 100.00 & 85.15 & 90.72 & 99.43 & 92.19 \\
        Digit 7     & 49.93    & 99.81    & 52.87     & \cellcolor{blue!25} 96.92 & 99.79 & \cellcolor{blue!25} 95.45 & 78.16 & 99.63 & 78.95 \\
        Digit 8     & 59.08    & 99.81    & 61.63     & 78.67 & \cellcolor{blue!25} 100.00 & 80.29 & 71.89 & \cellcolor{blue!25} 100.00 & 72.69 \\
        Digit 9     & 57.59    & 99.31    & 60.23     & 93.71 & 99.67 & 91.99 & \cellcolor{blue!25} 93.06 & 99.34 & \cellcolor{blue!25} 93.61 \\
        2 and 7     & 32.53    & \cellcolor{orange!25} 98.66    & 33.42     & 55.18 & 99.83 & 56.31 & 75.05 & 99.54 & 74.74 \\
        9 and 6     & 36.38    & 99.85    & 36.27     & 47.27 & 99.57 & 47.70 & 80.97 & 99.28 & 79.66 \\
        Line        & \cellcolor{orange!25} 28.53    & \cellcolor{blue!25} 100.00    & \cellcolor{orange!25} 28.73     & 57.42 & \cellcolor{blue!25} 100.00 & 59.27 & 89.13 & 99.68 & 89.75 \\
        Circle      & 28.60    &  99.03    & 30.16     & \cellcolor{orange!25} 38.97 & 99.74 & \cellcolor{orange!25} 39.13 & \cellcolor{orange!25} 51.51 & 99.79 & \cellcolor{orange!25} 50.25 \\
        \midrule
        Average     & 58.73    & 99.63    & 60.30     & 75.82 & 99.84 & 76.41 & 81.56 & 99.55 & 82.26 \\
        \bottomrule
    \end{tabular}
    \label{tab:rq2-MNIST}}\\
    \subfloat[L-DNN1 and L-DNN2.]{   
    \scriptsize
    \begin{tabular}{l | r r r | r r r}
        \toprule
        \multirow{2}{1.3cm}{\textbf{Feature}} & \multicolumn{3}{c|}{\textbf{L-DNN1}} & \multicolumn{3}{c}{\textbf{L-DNN2}} \\
        \cmidrule{2-7}
        & $\mathbf{R_{tr}}$ & $\mathbf{P_{te}}$ & $\mathbf{R_{te}}$ & $\mathbf{R_{tr}}$ & $\mathbf{P_{te}}$ & $\mathbf{R_{te}}$ \\
        \midrule
        CLL         & \cellcolor{blue!25} 66.77 & 99.12 & \cellcolor{blue!25} 63.75 & 94.03 & 99.57 & 97.86 \\
        FL          & 62.98 & 99.07 & 58.06 & 89.31 & 99.78 & 89.24 \\
        MCL         & 60.41 & 99.94 & 60.46 & \cellcolor{blue!25} 99.94 & \cellcolor{blue!25} 99.90 & \cellcolor{blue!25} 99.98 \\
        CLL \& FL   & 35.49 & \cellcolor{orange!25} 99.06 & 32.78 & \cellcolor{orange!25} 63.88 & \cellcolor{orange!25} 99.10 & \cellcolor{orange!25} 58.83 \\
        MCL \& FL   & 34.87 & 99.10 & 32.93 & 64.97 & 99.72 & 72.35 \\
        CLL \& MCL  & \cellcolor{orange!25} 28.93 & \cellcolor{blue!25} 99.95 & \cellcolor{orange!25} 28.50 & 78.36 & 99.74 & 85.60 \\
        \midrule
        Average     & 48.24 & 99.37 & 46.08 & 81.75 & 99.64 & 83.98 \\
        \bottomrule
    \end{tabular}
    \label{tab:rq2-lymphoma}}
\end{table*}

\emph{Results.} \Cref{sec:TableComparison} reports the recall of the rules on the training ($R_{tr}$) and the test ($R_{te}$) dataset, and their precision on the test dataset ($P_{te}$).
The maximum and minimum values for recall and precision have a blue and orange background.

\emph{Train Recall.} 
For M-DNN1, the rules for features ``Digit 1'' and ``Line'' have the highest (89.47\%) and lowest (28.53\%) train recall.
For M-DNN2, the rules for features ``Digit 7'' and ``Circle'' have the highest (96.92\%) and lowest (38.97\%) train recall.
For M-DNN3, the rules for features ``Digit 9'' and ``Circle'' have the highest (93.06\%) and lowest (51.51\%) train recall.
M-DNN1, M-DNN2, and M-DNN3 have an average train recall of 58.73\%, 75.82\%, and 81.56\%.
Therefore, M-DNN3 produces (on average) results with a higher train recall than M-DNN1 (+22.83\%) and M-DNN2 (+5.74\%).

For L-DNN1, the rules for features ``CLL'' and ``CLL \& MCL'' have the highest (66.77\%) and lowest (28.93\%) train recall.
For L-DNN2, the rules for features ``MCL'' and ``CLL \& FL'' have the highest (99.94\%) and lowest (63.88\%) train recall.
On average, the train recall for L-DNN2 (81.75\%) is significantly higher (+33.51\%) than the baseline L-DNN1 (48.24\%).

\emph{Test Precision.}
For M-DNN1, the rules for features ``Digit 0'', ``Digit 5'', and ``Line'' have the highest (100\%) test precision. 
The rule for feature ``2 and 7'' has the lowest (98.66\%) test precision.
For M-DNN2, the rules for features ``Digit 1'', ``Digit 2'', ``Digit 3'', ``Digit 4'', ``Digit 6'', ``Digit 8'', and ``Line'' have the highest (100\%) test precision. The rule for feature ``Digit 0'' has the lowest (99.46\%) test precision.
For M-DNN3, the rules for features ``Digit 4'' and ``Digit 8'' have the highest (100\%) test precision.  
The rule for feature ``Digit 5'' has the lowest (99.08\%) test precision. 
The average test precision is 99.63\% for M-DNN1, 99.84\% for M-DNN2, and 99.55\% for M-DNN3.
Therefore, the test precision of the three networks is comparable.

For L-DNN1, the rules for features ``CLL \& MCL'' and ``CLL \& FL'' has the highest (99.95\% ) and lowest (99.06\%) test precision.
For L-DNN2, the rules for features ``MCL'' and ``CLL \& FL'' have the highest (99.90\%) and lowest (99.10\%) test precision. 
The average test precision is 99.37\% for L-DNN1 and 99.64\% for L-DNN2.
These results indicate that the introduction of dropout has a negligible effect on the high precision of the extracted rules.

\emph{Test Recall.}
For M-DNN1, the rules for features ``Digit 1'' and ``Line'' have the highest (90.17\%) and lowest (28.73\%) test recall.
For M-DNN2, the rules for features ``Digit 7'' and ``Circle'' have the highest (95.45\%) and lowest (39.13\%) test recall. 
For M-DNN3, the rules for features ``Digit 9'' and ``Circle'' have the highest (93.61\%) and lowest (50.25\%) test recall.
The average test recall is 60.30\%, 76.41\% and 82.26\% for M-DNN1, M-DNN2, and M-DNN3.
Therefore, M-DNN3 produces (on average) results with a higher test recall than M-DNN1 (+21.96\%) and M-DNN2 (+5.85\%).

For L-DNN1, the rules for features ``CLL'' and ``CLL \& MCL'' have the highest (63.75\%) and lowest (28.50\%) train recall.
For L-DNN2, the rules for features ``MCL'' and ``CLL \& FL'' have the highest (99.98\%) and lowest (58.83\%) train recall.
The average test recall for L-DNN2 is 83.98\%, which is a significant improvement of +37.90\% over the L-DNN1 average of 46.08\%.

\begin{Answer}[RQ2 --- Influence of the Neural-Network]
The test precision of the \fga is not significantly affected by the type of neural network.
Conversely, the choice of the network significantly influences the recall of \fga: The variation of the (train and test) recall across our three study subjects for \Mnist is (on average) between +5.74\% and +22.83\% (on average) and between +33.51\% and +37.90\% for \Lymphoma.
\end{Answer} 
\subsection{Influence of the Composition of the Training Dataset  (RQ3)}
\label{sec:training}
To assess how the composition of the training dataset for the neural network influences the effectiveness of \fga, we proceeded as follows.

\emph{Methodology.} 
We performed $k$-fold cross-validation~\cite{mclachlan2005analyzing} on M-DNN3 and L-DNN2.
We selected M-DNN3 and L-DNN2 since they show the highest train and test recall on \Mnist and \Lymphoma (see \Cref{sec:nncomparison}).
For M-DNN3, we considered $k$ equal to $7$ to maintain the same ratio (6 to 1) between training and testing datasets as in \Mnist.
Note that for Experiment G, we have the same network used in \Cref{sec:nncomparison}.
For L-DNN2, we considered $k$ equal to $5$ since it was the value used for $k$-fold cross-validation in the original publication~\cite{janowczyk2016}.
Note that Experiment A refers to the network from \Cref{sec:nncomparison}.

\emph{Results.} 
\Cref{tab:rq3} reports the average recall of the rules on the training ($R_{tr}$) and the test ($R_{te}$) dataset, and their average precision on the test dataset ($P_{te}$) for M-DNN3 and L-DNN2.
It also reports the average value across all $k$ versions of the network and the difference between the maximum and minimum value (max2min).
The average is performed considering the top rule of every feature.
In this table, we adopt a background color blue for the highest value and orange for the lowest one.

\begin{table*}[t]
    \centering
    
    \caption{
    The average Train ($R_{tr}$) and Test ($R_{te}$) Recall and the Test Precision ($P_{te}$) of the rules extracted by the network for
    retraining in the k-fold cross-validation (Experiment).
    }
     \scriptsize
    \begin{tabular}{r | r r r | r r r }
        \toprule
        \multirow{2}{20mm}{\textbf{Experiment}}  &\multicolumn{3}{c|}{\textbf{M-DNN3}}   &\multicolumn{3}{c}{\textbf{L-DNN2}}\\
        \cmidrule{2-7}
        & $\mathbf{R_{tr}}$    & $\mathbf{P_{te}}$  & $\mathbf{R_{te}}$ & $\mathbf{R_{tr}}$    & $\mathbf{P_{te}}$    & $\mathbf{R_{te}}$\\
        \midrule
        A   & \cellcolor{blue!25} 84.96 & 99.52 & \cellcolor{blue!25} 85.36 & 81.75 & 99.64 & 83.98 \\
        B   & 80.40 & 99.47 & 78.99 & 80.87 & \cellcolor{blue!25} 99.73 & 80.05 \\
        C   & 79.88 & 99.49 & 79.33 & 82.25 & 99.23 & 83.79 \\
        D   & 78.19 & \cellcolor{orange!25} 99.32 & 77.53 & \cellcolor{orange!25} 65.91 & 97.52 & \cellcolor{orange!25} 67.43 \\
        E   & 82.64 & 99.38 & 81.92 & \cellcolor{blue!25} 89.56 & \cellcolor{orange!25} 97.45 & \cellcolor{blue!25} 90.06 \\
        F   & \cellcolor{orange!25} 72.99 & 99.50 & \cellcolor{orange!25} 74.90 &   &   &   \\
        G   & 81.56 & \cellcolor{blue!25} 99.55 & 82.26 &   &   &   \\
        \midrule
        Average & 80.09  & 99.46  & 80.04  & 80.07   & 98.71   & 81.06  \\
        max2min & 11.97  & 0.23   & 10.46  & 23.65   & 2.28    & 22.63  \\
        \bottomrule
    \end{tabular}
    \label{tab:rq3}
\end{table*}

\emph{Train Recall.}
For M-DNN3, the highest and lowest train recall are obtained for Experiment A (84.96\%) and F (72.99\%).
The average across all the trained networks is 80.09\%, with a maximum difference of 11.97\% between the best and worst recall values.
For L-DNN2, the highest and lowest train recall are obtained for Experiment E (89.56\%) and D (65.91\%).
The average across all the trained networks is 80.07\%, with a maximum difference of 23.65\% between the best and worst recall values.
This result shows that the composition of the training dataset significantly influences the train recall (differences can reach 23.65\%).

\emph{Test Precision.}
For M-DNN3, the highest and lowest test precision are obtained for Experiment G (99.55\%) and D (99.32\%).
The average across all the trained networks is 99.46\%, with a maximum difference of 0.23\% between the best and worst training.
For L-DNN2, the highest and lowest test precision are obtained in Experiment B (99.73\%) and E (97.45\%).
The average across all trained networks is 98.71\%, with a maximum difference of 2.28\% between the best and the worst training.
This result shows that there is a moderate influence (differences can reach 2.28\%) of the composition of the training dataset on the test precision.

\emph{Test Recall.} 
For M-DNN3, the highest and lowest test recall are obtained for Experiment A (85.36\%) and F (74.90\%).
The average across all trained networks is 80.04\%, with a maximum difference of 10.46\% between the best and the worst training.
For L-DNN2, the highest and lowest test recall are obtained for Experiment E (90.06\%) and D (67.43\%).
The average across all trained networks is 81.06\%, with a maximum difference of 22.63\% between the best and the worst training.
This result shows that the composition of the dataset significantly influences the test recall (differences can reach 23.65\%).

\begin{Answer}[RQ3 --- Influence of the Dataset Composition]
The choice of the training and test dataset significantly influences the recall of \fga.
For our benchmark, the training recall is affected by up to 23.65\% and the test recall by up to 22.63\%.
\end{Answer}
 
\subsection{Influence of the Feature Selection (RQ4)}
\label{sec:featureSelection}
We assessed the impact of feature selection on \fga's effectiveness as follows. 

\emph{Methodology}. 
For \Mnist, we considered all the possible 375 combinations\footnote{These combinations include the ones from \Cref{tab:features}.} of digits made by two, three, and four digits as features.
We considered combinations up to four digits: The highest number of digits aggregated by our features (see \Cref{tab:features}).
For \Lymphoma, we considered combinations of one and two features.

We run \fga for every combination as explained in \Cref{sec:replicability,sec:nncomparison,sec:training}.
For \Mnist and \Lymphoma, we used the neural network M-DNN3 and L-DNN2 since they have the highest train and test recall among the neural networks compared in \Cref{sec:nncomparison}.
We run our experiment on a large computing platform\footnote{1109 nodes, 64 cores, memory 249G or 2057500M, CPU 2 x AMD Rome 7532 2.40 GHz 256M cache L3.} to account for the number of combinations of digits to be considered. This reduced the time to approximately nine hours.
Note that we had to retrain the M-DNN3 to be compatible with the version of TensorFlow available on the server (2.11), which differs from the one (2.13) used to run the previous experiments.

To assess the behavior of \fga on combinations of features, we proceeded as follows.
We sorted the extracted rules by train recall (in descending order) since it is the criterion used to select the best-performing rule (see \Cref{sec:replicability}).
Intuitively, the rules with the highest recall are the ones that can successfully identify the highest percentage of inputs showing the feature.

\emph{Results}. \Cref{tab:rq4-mnist} (top part) shows the best 10 combinations ordered by train recall. 
\Cref{tab:rq4-mnist} (bottom part) report the results from \Cref{tab:rq2-MNIST} related to the features based on visual similarities.
To increase the readability of our work, we also report the results obtained from the possible combinations of two among the three Lymphoma subtypes from \Cref{tab:rq2-lymphoma} in \Cref{tab:rq4-lymphoma}.

\emph{Train Recall}. For the best 10 combinations of \Mnist, the rules for the features ``(0, 6)'' and  ``(1, 6)'' have the highest (98.09\%) and lowest (94.29\%) train recall. 
For our original combinations, the rules for the features ``Line'' and ``Circle'' have the highest (93.42\%) and lowest (69.78\%) train recall. 
For \Lymphoma, `` CLL \& MCL'' and ``CLL \& FL'' have the highest (78.36\%) and lowest (63.88\%) train recall.
This result shows that the train recall changes significantly (28.31\% for \Mnist and 14.48\% for \Lymphoma) depending on the feature selection.

\emph{Test Precision}. For the best 10 combinations of \Mnist, the rules for the features ``(1, 2)'' and ``(4, 9)'' have the highest (99.56\%) and lowest (98.90\%) test precision. 
For our original combinations, the rules for the features ``Line'' and ``Circle'' have the highest (99.69\%) and lowest (99.06\%) test precision. 
For \Lymphoma, ``CLL \& MCL'' and ``CLL \& FL'' have the highest (99.74\%) and lowest (99.11\%) test precision.
This result indicates that feature selection has a negligible impact (less than 1\%) on the precision of the rules computed by FGA.

\emph{Test Recall}. For the best 10 combinations of \Mnist, the rules for the features ``(0, 6)'' and ``(1, 6)'' have the highest (97.91\%) and lowest (93.82\%) test recall.  
For our original combinations, the rules for the features ``Line'' and ``Circle'' have the highest (92.81\%) and lowest (70.73\%) test recall. 
For \Lymphoma, ``CLL \& MCL'' and ``CLL \& FL'' have the highest (85.60\%) and lowest (58.83\%) test recall.
This result shows that the train recall changes significantly (27.18\% for \Mnist and 26.77\% for \Lymphoma) depending on the feature selection.\\

\begin{table*}[t]
    \centering
    \caption{The Train ($R_{tr}$) and Test ($R_{te}$) Recall and the Test Precision ($P_{te}$) of the 10 combinations of digits (Digit) with the highest Train Recall.}
    \label{tab:rq4}
    \hfill
    \subfloat[\Mnist dataset]{
     \scriptsize
    \begin{tabular}{l | r r r }
        \toprule
        \textbf{Digits}&$\mathbf{R_{tr}}$    & $\mathbf{P_{te}}$  &$\mathbf{R_{te}}$\\
        \midrule
         (0, 6)  &  \cellcolor{blue!25} 98.09 & 99.37 & \cellcolor{blue!25} 97.91\\
         (4, 9)     & 96.72    & \cellcolor{orange!25} 98.90 & 96.18\\
         (0, 5)     & 95.59    & 99.11 & 96.07\\
         (1, 7)     & 95.39    & 99.32 & 94.69\\
         (0, 2, 6)     & 95.09    & 99.29 & 95.58\\
         (1, 2)     & 94.90    &\cellcolor{blue!25} 99.56 & 94.98\\
         (1, 2, 7)     & 94.87    & 99.34 & 94.82\\
         (2, 6)     & 94.85    & 99.36 & 95.57\\
         (0, 2)     & 94.73    & 99.42 & 94.60\\
         (1, 6)     &\cellcolor{orange!25} 94.29    & 99.54 & \cellcolor{orange!25}93.82\\
         \midrule
         2 and 7 & 92.53 & 99.32 & 92.55\\
         9 and 6 & 87.94 & 99.18 & 87.78\\
         Line & \cellcolor{blue!25}93.42& \cellcolor{blue!25} 99.69 & \cellcolor{blue!25}92.81\\
         Circle & \cellcolor{orange!25} 69.78 & \cellcolor{orange!25}99.06 & \cellcolor{orange!25} 70.73\\
        \bottomrule
    \end{tabular}
    \label{tab:rq4-mnist}
    }
    \hfill
    \subfloat[\Lymphoma dataset]{
     \scriptsize
    \begin{tabular}{l | r r r }
        \toprule
        \textbf{Lymphoma Subtypes}&$\mathbf{R_{tr}}$    & $\mathbf{P_{te}}$  &$\mathbf{R_{te}}$\\
        \midrule
CLL \& FL   & \cellcolor{orange!25} 63.88     & \cellcolor{orange!25} 99.10     & \cellcolor{orange!25} 58.83 \\
        MCL \& FL   & 64.97     & 99.72     & 72.35 \\
        CLL \& MCL  & \cellcolor{blue!25} 78.36     & \cellcolor{blue!25} 99.74     & \cellcolor{blue!25} 85.60 \\
        \bottomrule
    \end{tabular}
    \label{tab:rq4-lymphoma}
    }
    \hfill
\end{table*}

\begin{Answer}[RQ4 --- Influence of the Feature Selection]
The feature selection has negligible influence (less than $1\%$) on the test precision of the rules computed by \fga.
However, the feature selection significantly affects the \fga recall with differences that reached $28.31\%$ for train recall and $27.18\%$ for test recall.
\end{Answer}

 \section{Discussion and Threats to Validity}
\label{sec:discussion}
We provide reflections and discuss threats to validity.

\emph{Discussion}. The results from RQ1 show that the precision of FGA is higher than that previously reported in the research literature. 
Practitioners working in domains where the precision of the techniques is of primary importance (e.g., safety-critical systems) may benefit from these results, which enrich and support a broader applicability of FGA.
These results also benefit the research community: The authors of FGA and other researchers working on similar techniques can benefit from our results.
Unlike the original work, our study provides a complete replication package that enables the reproducibility of our experiments.

The results from RQ2 and RQ3 show that the neural network selection and its training significantly affect the recall of the rules computed by FGA, while it does not significantly influence their precision. 
This result is relevant for practical applications.
First, the results sustain the use of FGA in domains where it is of primary importance that the feature is present when the rules are activated (precision) and it is acceptable that these rules do not cover some data possessing certain features (recall). 
Second, the results suggest industries can evaluate different types of neural networks
to increase the recall, since the neural network selection and training are primary factors that influence the recall of the rules.

Our results also suggest that selecting pure nodes (see \Cref{sec:background}) is effective in computing rules with high precision. 
However, this decision does not ensure a high recall.
Future research results should try to mitigate this drawback and develop techniques that can increase the recall of the rules produced by FGA.

The results from RQ4 suggest that neural network reasoning does not follow human reasoning.
For example, RQ4 shows that none of the features we defined considering visual similarities are among the best 10 combinations. 
This result indicates that FGA can compute rules that identify combinations of digits not visually similar to humans.
For example, the results from \Cref{tab:rq4} specify that the rule that identifies ``1'' and ``6'' is more effective than the rule that identifies ``9'' and ``6''.
We expected an opposite result, considering visual similarities.

FGA is effective in detecting a rule for the feature related to the digits ``0'' and ``6'', i.e.,  the feature ``(0, 6)'' has the highest (98.09\%) train recall in \Cref{tab:rq4}. 
This result is not surprising as both digits display a circular shape. 
The best ten rules computed by FGA also contain features representing combinations of digits that we did not consider similar. 
For example, the rule for the feature ``(1, 6)'' is among the 10 highest rules by train recall (94.29\%). 
We inspected some of the images to understand this behavior
and deduced that the handwriting style significantly affects the similarity of the digits ``0'' and ``6'' and ``1'' and ``6''.

Among the best 10 rules computed by FGA, there are also rules for features aggregating three digits, i.e., ``(0, 2, 6)'' and ``(1, 2, 7)''. 
This result was unexpected: We expected commonalities among two digits to be easier to identify than commonalities with three digits.
It is also surprising that ``(1, 2, 7)'' is among the best ten rules, but ``(1, 2)'' is not.
We speculate that the digits ``1'', ``2'', and ``7'' may have several elements in common between all three so the decision tree may not be as successful in separating two digits from the third.

\emph{Threats to Validity}. The selection of our benchmark threatens the \emph{external validity} of our results. 
Unfortunately, the dataset for the \Taxinet network \cite{Beland_2020,Frew_2004} (from Boeing) is not publicly available. 
For \Yolo-Tiny, the nuImages of the dataset are publicly available, but the authors trained a custom network, and the parameters used are not specified~\cite{Wang_2021,yolov4}.
The authors also did not disclose the 4000 images they considered among the 93000 images from the nuImages dataset. 
Since our results would not have been comparable, we considered two other datasets, since this choice provides more relevant results to understand the applicability of FGA in different domains.
The fact that for MNIST and \Lymphoma we used more images (70000 and 442398 images) than the ones the authors used for \Taxinet (450 images) and \Yolo-Tiny (4000 images) mitigates this threat.

The definition of the features from our experiments threatens the \emph{internal validity} of our results since considering other features may lead to different results.
However, the features containing multiple digits sharing similar characteristics were also considered by Crabb{\'e} et al.~\cite{crabbe2022} (i.e., ``vertical line'' for digits 1,4,7, as well as ``loop'' in digits 0,2,6,8,9), another approach extracting visual concepts from the input images of the \Mnist dataset. 
Unfortunately, a direct comparison between FGA and the approach from Crabb{\'e} et al.~\cite{crabbe2022} is meaningless since the rules computed by the two approaches are different and not comparable.

 \section{Related Work}
\label{sec:related}
We report on related work that: (a) considers the replicability, reproducibility, and repeatability of experiments in the machine learning field, and (b) approaches that focus on analyzing neural network models. 

\emph{Replicability, Reproducibility, and Repeatability (RRR)}. RRR of experiments is widely recognized as pivotal for the machine learning and software engineering domains~\cite{alahmari2020challenges,gonzalez2023revisiting,semmelrock2025reproducibility}.
The improper documentation of the experiments and limited access to the software code and data are recognized as challenges for the RRR activities~\cite{semmelrock2025reproducibility}.
For this reason, the research community is increasingly working on replicating machine learning experiments (e.g.,~\cite{arrieta2023deep,DBLP:journals/ese/DellAnnaAD23,DBLP:journals/ese/GiamatteiBPRT25,hassani2025empirical,kang2023large,liang2024can,melchor2022model,sabetzadeh2025practical,shah2025towards,wu2024reality}).
Unlike other replication studies (e.g.,~\cite{DBLP:journals/ese/DellAnnaAD23,DBLP:journals/ese/GiamatteiBPRT25,kang2023large}), we consider FGA.
Our results showed that FGA has higher precision than the one from the literature.

\emph{Explainability of ML}.  Approaches that analyze machine learning models to explain their behaviors have been classified by existing surveys~\cite{saleem2022explaining,samek2021explaining}.
For example, many approaches try to explain predictions by associating them with input features~\cite{lundberg2017unified,ribeiro2016should,shrikumar2016not,sundararajan2017axiomatic}.
Bondarenko et al.~\cite{bondarenko2017classification} propose an approach that extracts knowledge from a sigmoidal neural network as a binary classification decision tree.
Kim et al.~\cite{Kim_2018} proposed Concept Activation Vectors (CAVs). 
CAVs provide an interpretation of a vector space representing the neural network's internal state and the input feature as a vector space representing human-friendly concepts.
Crabb\'e and van der Schaar~\cite{crabbe2022} extended the CAV concept by introducing 
concept activation regions (CAR), which can compute explanations also for non linearly separable concepts. 
Many approaches to localize faults in a neural network have been proposed in the literature~\cite{cao2022deepfd,duran2021blame,eniser2019deepfault,ghanbari2023mutation,lyu2025fault,ma2018mode,schoop2021umlaut,sohn2023arachne,wardat2022deepdiagnosis,wardat2021deeplocalize}, which try to identify (and explain) the cause of the faults.

\fga~\cite{Gopinath_2023} is inspired by Prophecy~\cite{Gopinath_2019,gopinath2025prophecy}. 
Unlike \fga, Prophecy infers formal properties that (a)~ compute rules based on the activation status ``on''/``off'' of the neurons of the neural network and (b)~extract rules for correct vs missclassified inputs.
This problem differs from the one addressed by \fga.
Therefore, although Prophecy was also assessed on \Mnist, a comparison is meaningless since they target different problems.

In our work, we considered FGA among all these studies since it is a sound solution that has shown promising results in two industrial case studies. 
However, it is still not widely accepted and employed in the practical domain. 
Our results confirmed the maturity and potential practical benefit of this solution, supporting its wider application in practice.

 \section{Conclusion}
\label{sec:conclusion}
This paper assessed the applicability of FGA on a new benchmark made of the \Mnist and \Lymphoma datasets.  
We propose a new implementation of the \fga algorithm, since the original implementation is not publicly available.
We compared the results from our benchmark with those from the research literature.
Our results showed that \fga has higher precision on our benchmark than those from the literature.
Therefore, our results confirmed the maturity and the potential practical benefit of \fga, supporting a wider application of this technique in practice.
This paper also assessed how the selection of the neural network, training, and feature selection affect the effectiveness of the rules computed by \fga.
Our results showed that their selection significantly affects the recall of \fga, while it has a negligible impact on the precision of \fga.

We discussed the practical applications of our results. 
We also evidenced that our results suggest that selecting the pure nodes (see \Cref{sec:background}) enables the computation of rules with a high precision, but does not ensure a high recall.
Future research should try to mitigate this drawback and develop techniques that can increase the recall of the rules produced by FGA. 
\begin{credits}

\subsubsection{\discintname}
The authors have no competing interests to declare that are relevant to the content of this article.

\subsubsection{Data Availability}
A complete replicability package is available online~\cite{replication}.

\end{credits}

\bibliographystyle{splncs04}

\end{document}